\pdfoutput=1

\documentclass[11pt]{article}

\usepackage[]{acl}
\usepackage{graphicx}
\usepackage{booktabs}
\usepackage{soul,xcolor}

\usepackage{float}
\usepackage{times}
\usepackage{latexsym}

\usepackage[T1]{fontenc}

\usepackage[utf8]{inputenc}

\usepackage{microtype}

\title{A Keyword Based Approach to Understanding the Overpenalization of Marginalized Groups by English Marginal Abuse Models on Twitter}

\author{Kyra Yee \\
  Twitter \\
   \\\And
  Alice \\ 
  \textbf{Schoenauer Sebag} \\
  Twitter
   \\\And
 Olivia Redfield \thanks{\hspace*{0.5em} Work done while at Twitter}
 \\
  \\\And
  Emily Sheng \footnotemark[1]{} \\
  \\\And
  Matthias Eck \\
  Twitter
  \\\And
  Luca Belli \\ 
  Twitter
  }

\begin{document}
\setstcolor{red}

\maketitle
\begin{abstract}
{\it Content warning: contains references to offensive language}

Harmful content detection models tend to have higher false positive rates for content from marginalized groups. In the context of marginal abuse modeling on Twitter, such disproportionate penalization poses the risk of reduced visibility, where marginalized communities lose the opportunity to voice their opinion on the platform. Current approaches to algorithmic harm mitigation, and bias detection for NLP models are often very ad hoc and subject to human bias. We make two main contributions in this paper. First, we design a novel methodology, which provides a principled approach to detecting and measuring the severity of potential harms associated with a text-based model. Second, we apply our methodology to audit Twitter’s English marginal abuse model, which is used for removing amplification eligibility of marginally abusive content. Without utilizing demographic labels or dialect classifiers, we are still able to detect and measure the severity of issues related to the over-penalization of the speech of marginalized communities, such as the use of reclaimed speech, counterspeech, and identity related terms. In order to mitigate the associated harms, we experiment with adding additional true negative examples and
 find that doing so provides improvements to our fairness metrics without large degradations in model performance. 

\end{abstract}

\section{Introduction}

Because of the sheer volume of content, automatic content governance has been a crucial tool to avoid amplifying abusive content on Twitter. Harmful content detection models are used to reduce the amplification of harmful content online. These models are especially important to historically marginalized groups, who are more frequently the target of online harassment and hate speech \cite{amnesty2018toxic,vogels2021state}. However, previous research indicates that these models often have higher false positive rates for marginalized communities, such as the Black community, women, and the LGBTQ community  \cite{sap2019risk,oliva2021fighting,park2018reducing}. 
Within the context of social media, higher false positive rates for a specific subgroup pose the risk of reduced visibility, where the community loses the opportunity to voice their opinion on the platform. Unfortunately, there are many contributing factors to over-penalization, including linguistic variation, sampling bias, annotator bias, label subjectivity, and modeling decisions \cite{park2018reducing, sap2019risk,wich2020impact,ball2021differential}. 
This type of over-penalization risks  hurting the very communities  content governance
 is meant to protect. Algorithmic audits have become an important tool to surface these types of problems. However, determining the proper subgroups for analysis in global settings, and collecting high quality demographic information can be extremely challenging and pose the risk of misuse \cite{andrus2021we,holstein2019improving}. 
Current approaches to harm mitigation are often reactive and subject to human bias \cite{holstein2019improving}.
In this work, we present a more principled and proactive approach to detecting and measuring the severity of potential harms associated with a text-based model, and conduct an audit of one of the English marginal abuse models used by Twitter for preventing potentially harmful out-of-network recommendations. We develop a list of keywords for evaluation by analyzing the text of previous false positives to understand trends in the model's errors. This allows us to alleviate concerns of false positive bias in content concerning or created by marginalized groups without using demographic data. 

\section{Related Work}

\subsection{Challenges in Algorithmic Auditing in Industry}
As issues of algorithmic bias have become more prominent, algorithmic auditing has received increasing attention both in academia and by industry practitioners \cite{yee2021image,raji2020closing,buolamwini2018gender}. However, substantial challenges still remain for successfully being able to proactively detect and mitigate problems:
\begin{enumerate}

\item {\it Determining the appropriate subgroups for bias analysis}: Although algorithmic auditing has become a crucial tool to uncover issues of bias in algorithmic systems, audits can often suffer major blindspots and fail to uncover crucial problems that are not caught until after deployment or public outcry \cite{shen2021everyday,holstein2019improving, yee2021image}. This is often due to limited positionality and cultural blindspots of the auditors involved, or sociotechnical considerations that are difficult to anticipate before the system is deployed \cite{shen2021everyday,holstein2019improving}. Current approaches to bias detection often rely on predetermining an axis of injustice and acquiring demographic data, or for NLP models, pre-defining a lexicon of terms that are relevant to different subgroups \cite{dixon2018measuring,ghosh2021detecting,sap2019risk}.
Without domain expertise and nuanced local cultural knowledge, it may be difficult to anticipate problems or to know what relevant categories or combinations of categories should be focused on \cite{andrus2021we,holstein2019improving}.  For products such as Twitter that have global reach, this problem is exacerbated due to the huge amount of cultural and demographic diversity globally, and "efforts to recruit more diverse teams may be helpful yet insufficient" \cite{holstein2019improving}. Even in cases where audits are conducted proactively, inquiries into problem areas are often subject to human bias. Biases in non-Western contexts are also frequently overlooked \cite{sambasivan2021re}.

\item {\it Sensitivity of demographic data}: Most metrics used to measure disparate impact of algorithmic systems rely on demographic information  \cite{barocas2017fairness,narayanan2018translation}. 
However, in industry settings, high quality demographic information can be difficult to procure \cite{andrus2021we}. 

Additionally, many scholars have called into question harms associated with the uncritical conceptualization of demographic traits such as gender, race, and disability \cite{hanna2020towards,keyes2018misgendering,hamidi2018gender,khan2021one,hu2020s,bennett2020point}. There are fundamental concerns that the use of demographic data poses the risk of naturalizing or essentializing socially constructed categories \cite{benthall2019racial, hanna2020towards,fields2014racecraft,keyes2018misgendering}. Lastly, in industry settings, clients or users may be uncomfortable with organizations collecting or inferring sensitive information about them due to misuse or privacy concerns \cite{andrus2021we}. Additionally, inferring demographic information may pose dignitary concerns or risks of stereotyping \cite{keyes2018misgendering,hamidi2018gender, andrus2021we}. Despite these risks and limitations, this is not to suggest that demographic data should never be used. Demographic data can certainly be appropriate and even necessary for addressing fairness related concerns in many cases. However, because of the challenges discussed here, there is increasing interest in developing strategies to detect and mitigate bias without demographic labels \cite{benthall2019racial,lazovich2022measuring,rios2020fuzze}. 
\end{enumerate}

\subsection{Bias in automated content governance}

One key challenge in quantifying bias in machine learning systems is the lack of a universal formalized notion of fairness; rather, different fairness metrics imply different normative values and have different appropriate use cases and limitations \cite{narayanan21fairness, barocas2017fairness}. For the purposes of this study, we are primarily concerned with {\it false positive bias} in marginal abuse modeling. Previous research indicates that models used to detect harmful content often have higher false positive rates for content about and produced by marginalized groups. Previous work has demonstrated this can happen for several reasons.  Because they appear  more frequently in abusive comments than non-abusive ones, identity terms such as "muslim" and "gay", as well as terms associated with disability \cite{hutchinson2020social}, and gender \cite{park2018reducing,borkan2019nuanced}, exhibit false positive bias \cite{dixon2018measuring,borkan2019nuanced}. Research also indicates that annotator bias against content written in AAVE (African-American Vernacular English) is also likely a contributing factor to model bias against the Black community. \cite{sap2019risk,ball2021differential, halevy2021mitigating}. \citet{harris2022exploring} find evidence that the use of profanity and different word choice conventions are a stronger contributor to bias against AAVE than other grammatical features of AAVE.

Counterspeech \cite{haimson2021disproportionate} and reclaimed speech \cite{halevy2021mitigating,sap2019risk} from marginalized communities are also commonly penalized by models.
In summary, false positive bias on social media is a type of representational harm, where both content concerning marginalized communities (in the case of counterspeech or identity terms) or produced by marginalized communities (in the case of dialect bias or reclaimed speech) receives less amplification than other content. This can also lead to downstream allocative harms, such as fewer impressions or followers for content creators.

Determining what counts as harmful is an inherently a subjective task, which poses challenges for equitable content governance.
 The operationalization of abstract theoretical constructs into observable properties is frequently the source of many fairness related harms \cite{jacobs2021measurement}. Annotators' country of origin \cite{salminen2018online}, socio-demographic traits \cite{prabhakaran2021releasing,goyal2022your}, political views \cite{waseem2016you} and lived experiences \cite{waseem2016you, prabhakaran2021releasing} can affect their interpretations. 
Hate speech annotations have notoriously low inter-annotator agreement, suggesting that increasing the quality and detail of annotation guidelines is crucial for improving predictions \cite{ross2017measuring}. This problem is exacerbated for borderline content, as inter-annotator agreement tends to be lower for content that that was deemed moderately hateful in comparison with content rated as more severely hateful \cite{salminen2019online}.

\section{Methodology}

\subsection{English marginal abuse modeling at Twitter}

While Twitter does remove content that violates rules on abusive behavior and hateful conduct, content that falls into the margins (known as "marginal abuse") often stays on the platform and risks posing harm to some users. 

Twitter uses a machine learning model for English to try to prevent marginally abusive content from being recommended to users who do not follow the author of such content.
The model is trained to predict whether or not a Tweet qualifies as one of the following content types \footnote{While they are collected, labels from the following categories are not subject to de-amplification: {\it allegation of criminal behavior, claims of moral inferiority,  advocates for other consequences,} and {\it other insult}}: {\it advocate for violence, dehumanization  or incitement of fear, sexual harassment, allegation of criminal behavior, advocates for other consequences (e.g., job loss or imprisonment),  malicious cursing/profanity/slurs, claims of mental inferiority, claims of moral inferiority, other insult}.

\begin{table}
\scalebox{0.8}{
\begin{tabular}{llrr}
\toprule
      training set  & abusive    & non-abusive &  overall \\
\midrule
FDR &  39,018 &  89,050 &   128,068 \\
  prevalence &8,175 &  378,415 &   386,590 \\
  baseline model total & 47,193 & 467,465 & 514,658 \\
\midrule
  mitigation sample & 7,987 & 36,039 & 46,414 \\
  mitigated model total & 55,180 & 503,504 & 561,072 \\
 \midrule
 Test set (table~\ref{table:basemetrics}) & 916 & 20,770 & 21,686 \\
\bottomrule
\end{tabular}
}

\caption{Size of the training data for the baseline model and mitigated model, split by sampling type. The baseline model is trained only on the FDR and prevalence samples, whereas the mitigated model also includes the mitigation sample.}
\label{table:baselinetraining}
\end{table}

Twitter regularly samples Tweets in English to be reviewed by human annotators for whether or not they fall into one of the content categories listed above, and these annotations are used as ground-truth labels to train the marginal abuse model. Each Tweet sampled for human annotation is reviewed by 5 separate annotators and the majority vote label is used.
The training and evaluation data Twitter uses for the marginal abuse model is primarily sampled via two mechanisms: FDR (false discovery rate) sampling and prevalence based sampling. Prevalence based sampling is random sampling based on a weighting from how many times the tweet was viewed, and is generally used to measure the prevalence of marginally abusive content being viewed on the platform. In contrast, FDR sampling is sampling Tweets that have a high predicted marginal abuse score (using the current marginal abuse model in production) or high probability of being reported. This helps collect marginally abusive examples since they are relatively sparse, compared to other content categories. The model is trained on prevalence and FDR data sampled from  April 29 2021 to September 27 2021. In figure \ref{table:baselinetraining}, we give the size of the training data for the baseline and mitigated model split by sampling mechanism. Samples are collected from all publicly available Tweets identified as being written in English.

The marginal abuse model outputs a continuous score between 0 and 1, where scores closer to 1 indicate a higher probability of being marginally abusive (falling into one of the content types outlined above). The model has approximately 100 million parameters, is trained using TensorFlow 2.5, and and takes less than six hours to train using 2 gpus. All Tweets detected as being in English across all countries are scored using the marginal abuse model. Twitter sometimes inserts content into someone's home timeline from someone that the user does not explicitly follow, which is referred to as out-of-network content.\footnote{Examples of out of network content include suggested topics, as well as showing users content someone they follow liked. See \url{https://help.twitter.com/en/using-twitter/twitter-timeline} for additional details.} Tweets with a score greater than a tuned threshold are removed as candidates for out-of-network injections. Model scores are also used to help identify when to prompt users who are about to post harmful content with an opportunity to pause and reconsider their Tweet \cite{katsaros2022reconsidering} and to help rank replies on the conversations page. In summary, the model is only used for deamplification, and is not used to remove content. \footnote{ Tweets are only removed when they are identified as violating the Twitter rules, \url{https://help.twitter.com/en/rules-and-policies/twitter-rules} and the marginal abuse model is not involved in this process.}

In Part 1, we analyze the model's errors in order to figure out what sort of content gets over-penalized by Twitter's marginal abuse model, and develop a more comprehensive list of keywords in a more principled fashion. In Part 2, we quantify the severity of over-penalization and measure the effectiveness of a simple data augmentation technique to mitigate bias \cite{borkan2019nuanced}. 

\subsection{Part 1: What types of content are being over-penalized by the English marginal abuse model?}
We select all English annotated Tweets from both FDR and prevalence sampling between April 1, 2021 to August 30, 2021 \footnote{The size of the data used for evaluation for each keyword is given in the appendix} (after the model training window) and their scores. We group Tweets into four categories: FP (false positive), FN (false negative), TP (true positive), TN (true negative).

We leverage the threshold used for filtering tweets from being considered as a candidate for out of network injection, and convert the scores from Twitter's marginal abuse model to imputed binary labels. In order to split the data into FP, FN, TP, TN, we compare these predicted binary labels and the labels provided by human annotators. We then train a linear model on top of a tf-idf (term frequency–inverse document frequency) representation of the Tweet to predict whether a given Tweet is misclassified as a FP by the marginal abuse model or not in comparison to the human annotated label. In other words, the linear model predicts a binary label for  FP vs. (TP, FN, TN) given the tf-idf representation of the Tweet. 
The tf-idf vector representation was learned using using \citet{scikit-learn}'s TfidfVectorizer on the entire corpus of annotated Tweets described above, where each Tweet was treated as a separate document. We perform stopword filtering using \citet{scikit-learn}'s English stopword list. Additionally, the vocabulary for the tf-idf vector representations ignores words that have a frequency above a specific threshold to get rid of corpus specific stopwords, as well as ignores words that have a frequency lower than a given threshold to avoid sparsity issues. We manually tune both these parameters, the final values used in the analysis are max\_df=0.05, min\_df=0.0002.
Since each feature of the linear model corresponds to a word in the vocabulary, we look at the heaviest weighted features to look for trends in the type of content that is over penalized with respect to the human annotations. The resulting tf-idf vocabulary has 6,313 words, and we look at the top 350 words, corresponding to approximately the top 5\% heaviest weighted features.  We manually group together some of the patterns observed within the top 350 coefficients. We manually aggregate plurals.

\subsection{Part 2: Measuring the severity of over-penalization and effectiveness of data augmentation for mitigation}
In Part 1, we developed a new technique to acquire a more holistic picture of areas of concern within the model's false positive predictions. Next, we would like to use more established metrics to measure the severity of bias and measure the effectiveness of a simple data augmentation strategy to attempt to mitigate the observed bias in the model. 
\subsubsection{Metrics Definitions}
For a given keyword, the metrics compare all Tweets containing that keyword, which is referred to as the {\it subgroup}, to the rest of the data, which is referred to as the { \it background}. We use the following metrics, see \citet{borkan2019nuanced} for details.

\begin{itemize}
\item {\it Subgroup AUC:} AUC measured on the subgroup of interest. This represents model understanding and separability for a given subgroup.
\item {\it Background Positive Subgroup Negative (BPSN) AUC:} 
AUC on the positive examples from the background and the negative examples from the subgroup. Lower scores would likely result in false positives for this subgroup at many thresholds. 

\item {\it Background Negative Subgroup Positive (BNSP) AUC:}  AUC on the negative examples from the background and positive examples from the subgroup. Lower scores would likely result in false negatives for this subgroup at many thresholds.

\end{itemize}
For all the AUC metrics, values closer to 1 are better and indicate a reduction in errors.\footnote{Per the suggestions in \citet{borkan2019nuanced} we also experimented with using AEG and NAEG. we found NAEG to be highly correlated with BPSN AUC for our keywords, which is probably due to the way we sampled our keywords for evaluation. For our data, confidence intervals for AEG seemed to be so large that the metric did not seem to provide much additional useful information beyond what is reflected in the AUC metrics}
95\% confidence intervals are computed using an empirical bootstrap.  Similar to Part 1, we evaluate on prevalence based and FDR based samples for all English Tweets globally, but sampled from February 2, 2022 to May 4, 2022.\footnote{Part 1 was conducted using an earlier version of the annotation task and model, and thus uses an earlier date range to evaluate. In industry settings it is sometimes unavoidable that the underlying model may be changed during the course of an audit. However, it is reasonable to assume problem areas from the earlier version should also be evaluated in the most recent version of the model/annotation task.} 
Given that we have a large number of keywords to evaluate on, we also employ the meta-metrics introduced by \citet{metametrics} in order to summarize differences in performance across subgroups and allow for more interpretable comparison between models. We look at two meta-metrics: 1. the standard deviation of group-wise model performance metrics, adjusted for each group's sample variance (var), and 2. the difference between the maximum and minimum group performance metrics (maxmin). For the meta-metrics, values closer to 0 are better as they represent a lower disparity between groups. The size of the data per keyword is given in the appendix in Table \ref{table:evalsize}.

\section{Results}
\subsection{Part 1: What types of content are being over-penalized by the English marginal abuse model?}
Below we have organized the terms by theme, and manually aggregated singular/plurals. Terms were selected from the top 350 coefficients for each word (approx top 5 percentile of scores).  We use the term ``identity related'' terms to discuss terms that relate to group identity and demographics, which we consider distinct from political identity for the sake of this analysis. We manually group together some of the patterns observed within the coefficients.
\begin{itemize}
\item {\bf identity related terms already included in the jigsaw evaluation set} \cite{borkan2019nuanced}: gay, white, muslim, jewish, trans, lesbian, female, male, black, queer, transgender, lgbtq, lgbt, american, chinese, deaf
\item {\bf new identity related terms}: islam, man, jew, gender, woman, muslim, religion,  POC, girl
\item {\bf reclaimed speech}: n****, bitch, fat, hoe, whore, ho, slut
 \item {\bf potential counterspeech}: racist, privileged
\item {\bf countries/regions}: america, palestine, russia, africa, ethiopia, afghanistan, china
\item {\bf political identity}: democrat, dem, republican, liberal,  libs, conservative, feminist, socialist, marxist, tory, communism, commie, communist, leftist, tories, progressives
\item {\bf political topics}: trump, biden, obama, markgarretson, hitler, cuomo, politicians, gop, trudeau, kamala, boris, cia, vote, clinton , fascism, atrocities, maga, fox, antifa, cheney, political, constituents
\item {\bf sexual terms}: ass, pussy, dick, cock, penis, cum, sex, arse, virgin, lick, bum, nuts, fucked
\item {\bf terms in grammatical constructions}: ass, fuck, fuckin
\item {\bf current events and topics of discussion}: taliban, terrorists, CIA, abortion, CRT, cop, abortions, ethiopia, palestine
\end{itemize}

We observe that issues with false positives extend beyond identity related terms and also include other classes of content that have been sources of concern in content governance for marginalized communities, such as reclaimed speech and counterspeech
\cite{haimson2021disproportionate,dixon2018measuring,halevy2021mitigating}. Reclaimed speech refers to the process when slurs, which have been traditionally used to disparage a group, are re-appropriated by the community targeted by the slur \cite{croom2011slurs,ritchie2017social,nunberg2018social}. 
The goal of re-appropriation can be to change the connotation of the word to be neutral or even positive (ex. mainstream adoption of the word "queer"), but in other cases the intent can be to retain the stigma in an act of defiance (ex. ``dyke marches'' or ``slut walks'' to draw awareness to issues of stigma and discrimination) \cite{brontsema2004queer,nunberg2018social}.  
Re-appropriation can be leveraged to express in-group solidarity and shared history \cite{croom2011slurs,ritchie2017social} and ``mock impoliteness'' has been demonstrated to help LGBTQ people deal with hostility \cite{oliva2021fighting, murray1979art,jones2007drag,mckinnon2017building}.  

False positives can also include terms like "ass" or "fuck" used in grammatical constructions, that aren't necessarily intended pejoratively \cite{napoli2009grammatical}. 
For example in AAVE, the use of a possessive with "ass" forms a nominal construction \cite{halevy2021mitigating,collins2008aae}, such as in "I saw his ass at the store yesterday". Ensuring the marginal abuse model does not over-index on profanity may also be useful because hate groups often frequently avoid profanity to avoid simplistic detection and to appear respectable \cite{adl}.

Because the effectiveness of bias mitigation techniques varies greatly between dataset contexts \cite{ball2021differential}, we emphasize that this list of keywords is specific to English marginal abuse modeling on Twitter and likely does not generalize well for evaluation of marginal abuse models in other contexts. 

\subsection{Part 2: What is the severity of over-penalization and effectiveness of the mitigation?}

\subsubsection{Mitigation Description}

 \begin{figure*}
        \centering
        \includegraphics[scale=0.3]{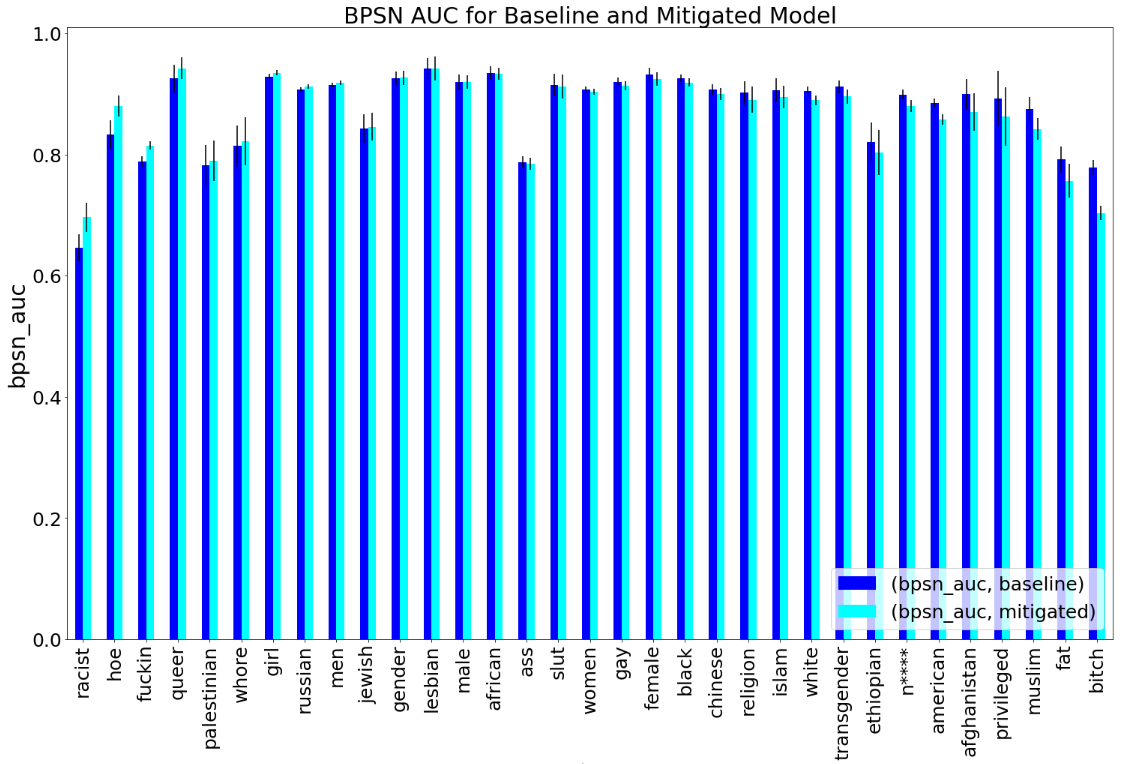}
        
        \caption{BPSN AUC for the baseline and mitigated model. BPSN AUC increases for some keywords and decreases for others. }
\label{table:bpsn}
    \end{figure*}

For a given keyword that occurs in both abusive and non-abusive settings, the current sampling mechanism (combining FDR sampling and prevalence sampling) oversamples abusive examples of Tweets containing the keyword in order to account for the general sparsity of abusive samples. Non-abusive Tweets with keywords are undersampled in comparison to their true distribution, so randomly sampling more Tweets with these keywords to acquire more true negatives could help reduce false positives and issues related to feedback loops in FDR sampling. This phenomenon was described for identity terms in marginal abuse models in \citet{dixon2018measuring}, but in this analysis we observe that this pattern is broadly generalizable to many classes of content. 

Given the analysis above and our focus on mitigating the risk of overpenalization of content related to and authored by historically marginalized groups, we restrict our mitigation to identity-related terms, reclaimed speech, counterspeech, countries/geographies, and grammatical intensifiers. Political bias and handling sexual content are left as an area of future work. 

For each of the keywords in the classes of content listed above, similar to \citet{dixon2018measuring}, we add additional random samples of Tweets containing a keyword to the training data in order to increase the number of true negative samples. For each keyword, the number of additional samples added was equal to 50\% of the number of non-abusive examples in the original training set, totaling approximately 46k new samples in total. We refer to this sample as the {\it mitigation sample}. The mitigation sample is drawn from February 1, 2022.  For hyperparameter tuning, the baseline model is retrained regularly and thus a set of reasonable hyperparameters was known. For the mitigated model, we ended up using this same set of hyperparameters since the training data largely overlaps with the baseline.

\subsubsection{Mitigation Results}

In Table~\ref{table:bpsn}, we present the difference in BPSN AUC for each of the selected keywords. 
We evaluate February 2, 2022 to May 11, 2022, a later date range than in Part 1 in order to only evaluate on samples drawn from after the mitigation sample.
We observe that the mitigation works inconsistently for different keywords, and is ineffective in significantly improving performance for many keywords. We conducted several additional experiments to try to determine why the mitigation works for some keywords and not others. We could not find any signal that could explain which keywords improve/degrade (see~\ref{sec:result_analysis} for details).

Because of the large number of subgroups we have, in Table \ref{table:metametrics} we also report results for the meta-metrics to better able to make human interpretable model comparisons. For all three of our underlying metrics (subgroup AUC, BPSN AUC, BNSP AUC), we observe improvements in both the variance and maxmin meta metrics \footnote{  \citet{metametrics} found that bootstrapped confidence intervals for meta metrics are statistically biased. A correction has been worked out for binary metrics, but not for AUC metrics. We therefore were unable to provide confidence intervals for our metrics at this time but consider this an important future area of work.
}. Therefore, we conclude the mitigated model is better than the baseline. In Table \ref{table:basemetrics}, we also look at the precision-recall (PR) and receiver operating characteristic (ROC) area under curves (AUC) as traditional measures of model performance. For these metrics, we look at a random sample of English tweets. This evaluation dataset is as close as possible to the underlying distribution of tweets on the platform, see appendix for details on evaluation set size.
For ROC AUC and PR AUC, we observe minor degredations to performance. In summary, we were able to demonstrate improvements to our fairness metrics without substantial degredations to overall model quality. 
However, fairness improvements are also minimal. Future directions include more advanced mitigation strategies,  as well as trying to understand why the mitigated tested here works inconsistently for different keywords.

\begin{table}
\begin{tabular}{llrr}
\toprule
         &     &  baseline &  mitigated \\
\midrule
subgroup\_auc & maxmin &  0.162 &   0.148 \\
         & var &  0.029 &   0.022 \\
bpsn\_auc & maxmin &  0.317 &   0.264 \\
         & var &  0.063 &   0.062 \\
bnsp\_auc & maxmin &  0.110 &   0.098 \\
         & var &  0.015 &   0.014 \\
\bottomrule
\end{tabular}
\caption{Meta-metrics comparing the mitigated and baseline model performance. The mitigated model demonstrates improvements in all meta-metrics, so we conclude the mitigated model is better than the baseline.}
\label{table:metametrics}
\end{table}

\begin{table}
\begin{tabular}{llrr}
\toprule
              &  baseline &  mitigated \\
\midrule
PR AUC &  0.657 (0.017) &   0.645 (0.017) \\
ROC AUC &   0.963 (0.003) &   0.961 (0.003) \\
\bottomrule
\end{tabular}
\caption{Aggregate model performance, comparing the mitigated and baseline models. Averages and standard deviations are provided over 100 bootstrap samples of the test set.}
\label{table:basemetrics}
\end{table}

\section{Limitations and Future Work}

This analysis relies on comparing model predictions with human annotations. One limitation of this approach is the following: we are assuming that the human annotated labels represent a reasonable ground truth. However, it’s likely that the annotations have their own bias issues. A future area of work is to analyze how reliable the annotations are for some of the top keywords surfaced here, especially for reclaimed speech and for Tweets with AAVE. However, because previous work has found that word choice and profanity are likely stronger contributors to bias against AAVE than linguistic features of AAVE \cite{harris2022exploring}, we hope that bias mitigation techniques at the keyword level can also help alleviate bias against AAVE without the use of sensitive racial or dialect classifiers. Another fruitful area of future work would be to better understand the relationship between mitigating bias at the keyword level versus the dialect level. 

Our methodology is helpful for detecting the most widespread and prevalent problems. However, there may be other serious problems that do not receive the same amount of traffic that still deserve attention. Oftentimes, smaller groups of people, especially those who live at the intersection of multiple marginalized identities can suffer the worst harms from algorithmic systems \cite{ajl}. Thus, relying on frameworks that focus on bigger segments of the population poses the risk of missing important harms to smaller communities. In this work, we develop a list of keywords for bias evaluation by analyzing a corpus generated from all English Tweets on Twitter. However, because English Twitter is primarily composed of users from the United States and the United Kingdom, our list of keywords for evaluation is likely heavily skewed towards US-centric or Western issues. One way to mitigate this would be to repeat the analysis conducted here, but using separate corpora for each country 
or upsampling Tweets from countries with smaller populations of Twitter users  in order to ensure we are getting appropriate coverage in other countries with smaller user bases. This would help increase coverage for minority groups in the data we use for bias evaluation.
 Another critical area of work would include expanding the analysis to other languages beyond English. The overemphasis of English has led to the underexposure of other languages in NLP research \cite{hovy2016social}.

This work treats reclaimed uses of slurs as an important facet of the speech of marginalized communities. However,
reclamation is not a "bullet-proof" process - some may find reappropriated uses acceptable and others may not.  Additionally, reclamation may only be deemed acceptable by in-group members or in certain contexts \cite{rahman2012n}. Since the marginal abuse model only uses the text of a single Tweet (and not any information about the Tweet author or conversational context), it is difficult for the model to account for such nuance. Furthermore, because this model is used to moderate all English content on Twitter, the model implicitly assumes the same utterance has the same meaning across the world, which is an extreme oversimplification. In other words, the model does not account for local variations in language use. Reclamation can also backfire, for example the Hong Kong media's mocking of the reclaimed use of "tongzhi" (literally meaning 'comrade') by the gay and lesbian community \cite{zimman2017transgender,wong2005reappropriation}. This example serves to illustrate the essentially contested nature of reclaimed speech and how language ideologies shift over time. With respect to automatic content governance, shifting language ideologies indicate the importance of 1) meaningfully engaging and consulting with affected communities on models used for content governance, 2) the utility of regular audits and model refreshes to account for change in language use over time, and 3) additional user controls to better accommodate for multiple definitions of harmful content. Lastly, there are inherent limitations to fixing socio-technical problems through purely technical means \cite{ball2021differential}. We hope that our analysis provides an interesting case study of some of the challenges associated with automatic content governance in industry and sparks further discussion.

\section{Conclusion}

Current approaches to harm mitigation and bias detection are frequently reactive and subject to human bias. Additionally, demographic labels and dialect classifier are difficult to acquire and pose ethical concerns in industry settings. In this paper, we present a novel approach for developing a list of keywords for bias evaluation of text based models in a more principled and proactive fashion. Looking at Twitter's English marginal abuse model, we are able to detect issues related to the over-penalization of speech concerning and produced by marginalized communities, such as reclaimed speech, counterspeech, and identity related terms without using demographic data. We demonstrate that a simple data augmentation mitigation is able to relieve some of the observed bias without causing substantial degradations in aggregate model quality. However, technical mitigation techniques are not a silver bullet. Due to the inherent subjectivity of marginal abuse, contested nature of reclaimed speech, and language change on social media, we emphasize the need for regularly conducted audits, additional user controls for content governance, and channels for community feedback for ML models used for content governance.

\bibliography{anthology,custom}
\bibliographystyle{acl_natbib}
\clearpage
\newpage
\appendix

\section{Appendix}
\label{sec:appendix}

\subsection{
Size of Evaluation Data per Keyword}
The size of the data used for evaluation for each keyword is given in table  \ref{table:evalsize}.

\begin{table}
\scalebox{0.8}{

\begin{tabular}{lrrr}
\toprule
keyword &  total count &  pos count &  neg count \\
\midrule
afghanistan &    197 &         49 &            148 \\
african     &    659 &        108 &            551 \\
american    &   3152 &        951 &           2201 \\
ass         &   2932 &       1209 &           1723 \\
bitch       &   2174 &       1060 &           1114 \\
black       &   1993 &        397 &           1596 \\
chinese     &   1134 &        175 &            959 \\
ethiopian   &    131 &         30 &            101 \\
fat         &    438 &        164 &            274 \\
female      &    547 &         80 &            467 \\
fuckin      &   7094 &       3220 &           3874 \\
gay         &   1353 &        234 &           1119 \\
gender      &    572 &         97 &            475 \\
girl        &   3545 &        490 &           3055 \\
hoe         &    435 &        178 &            257 \\
islam       &    261 &         52 &            209 \\
jewish      &    329 &         79 &            250 \\
lesbian     &    165 &         24 &            141 \\
male        &    516 &         87 &            429 \\
men         &   6843 &       1361 &           5482 \\
muslim      &    373 &         66 &            307 \\
n****       &   1284 &        351 &            933 \\
palestinian &    198 &         52 &            146 \\
privileged  &     79 &         24 &             55 \\
queer       &    173 &         38 &            135 \\
racist      &   1001 &        671 &            330 \\
religion    &    249 &         72 &            177 \\
russian     &   7082 &       1147 &           5935 \\
slut        &    244 &         56 &            188 \\
transgender &    631 &        111 &            520 \\
white       &   2143 &        554 &           1589 \\
whore       &    191 &         72 &            119 \\
women       &   4420 &        992 &           3428 \\
\bottomrule
\end{tabular}
}
\caption{Size of the evaluation data for each keyword for bias analysis}
\label{table:evalsize}

\end{table}
\subsection{Mitigation result analysis}
\label{sec:result_analysis}
As is visible on figure~\ref{table:bpsn}, the results from adding keyword-based samples to the training data did not consistently improve BPSN AUCs across keywords. We therefore tried multiple avenues of analysis to understand where the discrepancies could come from. First, we performed the same analysis grouping keywords in themes, and found similarly inconsistent results across the board.
Second, although we found larger standard deviation in BPSN AUCs results to be significantly correlated with smaller number of data points in the test set, we could not find any reason for the BPSN AUC values themselves.

In the following, correlation stands for Pearson correlation, and we used the same regular expressions to identify which Tweets contained which keywords in the test set, as had been done in the training set. Spearman correlations did not show any insight either and are not reported.

\subsubsection{Thematic analysis gives similarly inconsistent results to keyword analysis}
Following~\citet{borkan2019nuanced}, the keyword-based analysis relies on whether, for a given keyword, a Tweet contains it. If it does, it is included in the subgroup for that keyword, and if it doesn't, it is included in the background for that keyword. However, certain keywords belong to similar themes and are likely to occur in similar context (e.g. "bitch", "hoe", "slut" and "whore"). We therefore thought about grouping similar keywords into themes (e.g. "potentially insulting terms to describe a woman"). Although we are aware that such groupings are highly influenced by the background of whomever is making them, thematic groups are larger than keyword groups and have potentially less noisy backgrounds. There was therefore hope for more significant, and/or understandable, and/or consistent results.

\begin{figure*}
        \centering
        \includegraphics[scale=0.4]{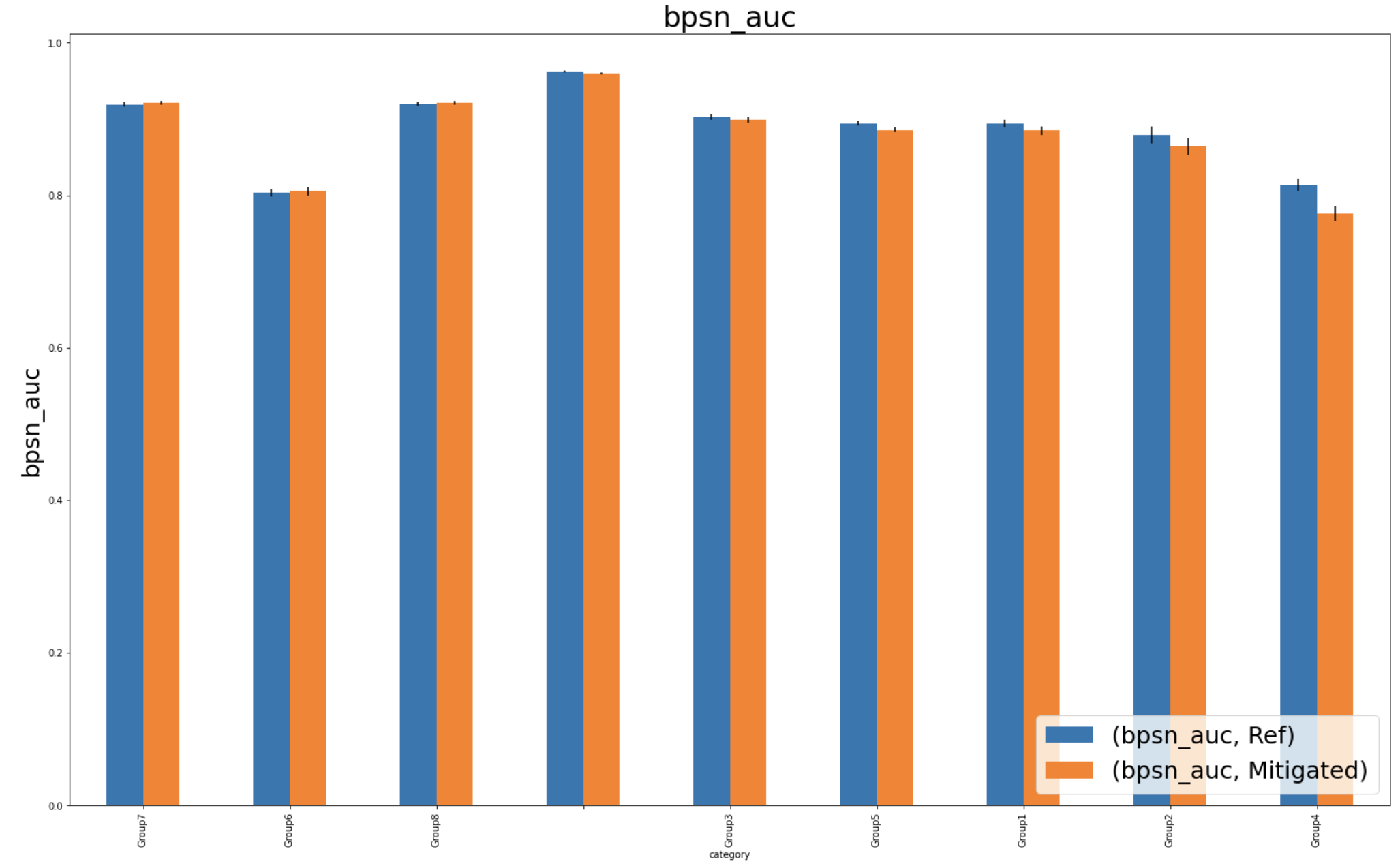}
        
        \caption{BPSN AUCs for the baseline and mitigated model. BPSN AUC increases for some themes and decreases for others. See text for theme descriptions. Confidence intervals are provided using 1000 bootstrap samples}
\label{fig:themes}
\end{figure*}

We manually designed eight groups: (1) "Race", (2) "Religion", (3) "National origin", (4) "Potentially insulting terms to describe a woman", (5) "Neutral and potentially insulting terms to describe a woman", (6) "Generally insulting terms", (7) "Gender", (8) "Gender and sexual orientation". We repeated the analysis as described in the main text, based on these thematic groupings of Tweets. Unfortunately, as can be seen on fig.~\ref{fig:themes}, certain groups did show an improvement in BPSN AUC between the baseline model, and the mitigated model (e.g. Group 7, "Gender"), while other groups did not (e.g. Group 4, "Potentially insulting terms to describe a woman"). In the following analyses, we include the thematic results next to the keyword results.

\subsection{BPSN AUCs standard deviations are negatively correlated with test set content}
Standard deviations in BPSN AUCs before and after training the marginal abuse model with the mitigated dataset are computed using bootstrap samples of the test set. The two sets of standard deviations are highly correlated (Pearson correlation, 0.985, p-value < 0.001). They are also highly correlated with the number of data points for each keyword in the test set, either only abusive or not. For example, the correlation between the standard deviation in mitigated BPSN AUCs and the number of data points for each keyword in the test set is -0.577 (p-value < 0.001).

This points to the fact that the test set itself should be sampled in a targeted fashion, to ensure being large enough with respect to rarer keywords.

\subsubsection{No data characteristic was found to be significantly linked to BPSN AUC changes}
We investigated the correlation between the difference in BPSN AUC, and the following characteristics of the dataset:
\begin{itemize}
    \item the number of datapoints, abusive or not, coming from the prevalence sample;
    \item the number of datapoints, abusive or not, coming from the FDR sample;
    \item the number of datapoints, abusive or not, coming from either the prevalence or the FDR sample;
    \item the number of datapoints, abusive or not, coming from the mitigation sample;
    \item the percentage of abusive datapoints coming either from the prevalence or the FDR sample;
    \item the percentage of abusive datapoints coming from the mitigation sample;
    \item the growth rate of the number of datapoints, abusive or not, between the training set of the baseline model and that of the mitigated model;
    \item the number of datapoints in the test set.
\end{itemize}
No correlation was significant (p-value > 0.1).
\end{document}